\documentclass[10pt,journal,compsoc]{IEEEtran}
\usepackage{lineno,hyperref}
\usepackage{times}
\usepackage{latexsym}
\usepackage{graphicx,subfigure}
\usepackage{epstopdf}
\usepackage{amsmath, amsfonts}
\usepackage{multirow}
\usepackage{upgreek}
\usepackage{bm}
\usepackage{cite}
\usepackage{indentfirst}
\usepackage{graphicx}
\usepackage{color}
\usepackage{amssymb}
\usepackage{algorithm}
\usepackage{algorithmic}
\usepackage{arydshln}

\begin{document}

\title{LSBert: A Simple Framework for Lexical Simplification}

\author{Jipeng~Qiang, Yun~Li, Yi~Zhu, Yunhao~Yuan, and~Xindong~Wu, ~\IEEEmembership{~Fellow,~IEEE,}
\IEEEcompsocitemizethanks{\IEEEcompsocthanksitem J. Qiang, Y. Li, Y. Zhu, and Y. Yuan are with the Department
of Computer Science, Yangzhou, Jiangsu, China.\protect\\
E-mail: \{jpqiang,liyun, zhuyi, yhyuan\}@yzu.edu.cn
\IEEEcompsocthanksitem X. Wu is with Key Laboratory of Knowledge Engineering with Big Data (Hefei University of Technology), Ministry of Education, Hefei, Anhui, China, and Mininglamp Academy of Sciences, Minininglamp, Beijing, China. \protect\\
E-mail: xwu@hfut.edu.cn
}
}

\markboth{Journal of \LaTeX\ Class Files,~Vol.~14, No.~8, April~2019}%
{Shell \MakeLowercase{\textit{et al.}}: Bare Advanced Demo of IEEEtran.cls for IEEE Computer Society Journals}

\IEEEtitleabstractindextext{%
\begin{abstract}

Lexical simplification (LS) aims to replace complex words in a given sentence with their simpler alternatives of equivalent meaning, to simplify the sentence. Recently unsupervised lexical simplification approaches only rely on the complex word itself regardless of the given sentence to generate candidate substitutions, which will inevitably produce a large number of spurious candidates. In this paper, we propose a lexical simplification framework LSBert based on pretrained representation model Bert, that is capable of (1) making use of the wider context when both detecting the words in need of simplification and generating substitue candidates, and (2) taking five high-quality features into account for ranking candidates, including Bert’s prediction order, Bert-based language model, and the paraphrase database PPDB, in addition to the word frequency and word similarity commonly used in other LS methods. We show that our system outputs lexical simplifications that are grammatically correct and semantically appropriate, and obtains obvious improvement compared with these baselines, outperforming the state-of-the-art by 29.8 Accuracy points on three well-known benchmarks.

\end{abstract}

}

\maketitle

\IEEEdisplaynontitleabstractindextext

\IEEEpeerreviewmaketitle

\section{Introduction}

Lexical Simplification (LS) aims at replacing complex words with simpler alternatives, which can help various groups of people, including children \cite{De_belder}, non-native speakers \cite{paetzold2016unsupervised}, people with cognitive disabilities \cite{feng2009automatic,saggion2017automatic}, to understand text better. LS is an effective way of simplifying a text because some work shows that those who are familiar with the vocabulary of a text can often understand its meaning even if the grammatical constructs used are confusing to them. The LS framework is commonly framed as a pipeline of three steps: complex word identification (CWI), substitute generation (SG) of complex words, and filtering and substitute ranking (SR). CWI is often treated as an independent task \cite{paetzold2017lexical}. Existing LS systems mainly focused on the two steps (SG and SR) \cite{gooding2019recursive}. 

The popular LS systems still predominantly use a set of rules for substituting complex words with their frequent synonyms from carefully handcrafted databases (e.g., WordNet) \cite{devlin1998the} or automatically induced from comparable corpora \cite{De_belder} or paraphrase database \cite{pavlick2016simple}. Recent work utilizes word embedding models to extract substitute candidates for complex words. Given a complex word, they extracted the top 10 words as substitute candidates from the word embedding model whose vectors are closer in terms of cosine similarity with the complex word \cite{glavavs2015simplifying,paetzold2016unsupervised,paetzold2017lexical}. Recently, the LS system REC-LS attempts to generate substitute candidates by combining linguistic databases and word embedding models. However, they generated substitute candidates for the complex word regardless of the context of the complex word, which will inevitably produce a large number of spurious candidates that confuse the systems employed in the subsequent steps. For example, if simpler alternatives of the complex word do not exist in substitute candidates, the filtering and substitute ranking step of LS is meaningless.

\begin{figure}
  \centering
  \includegraphics[width=75mm]{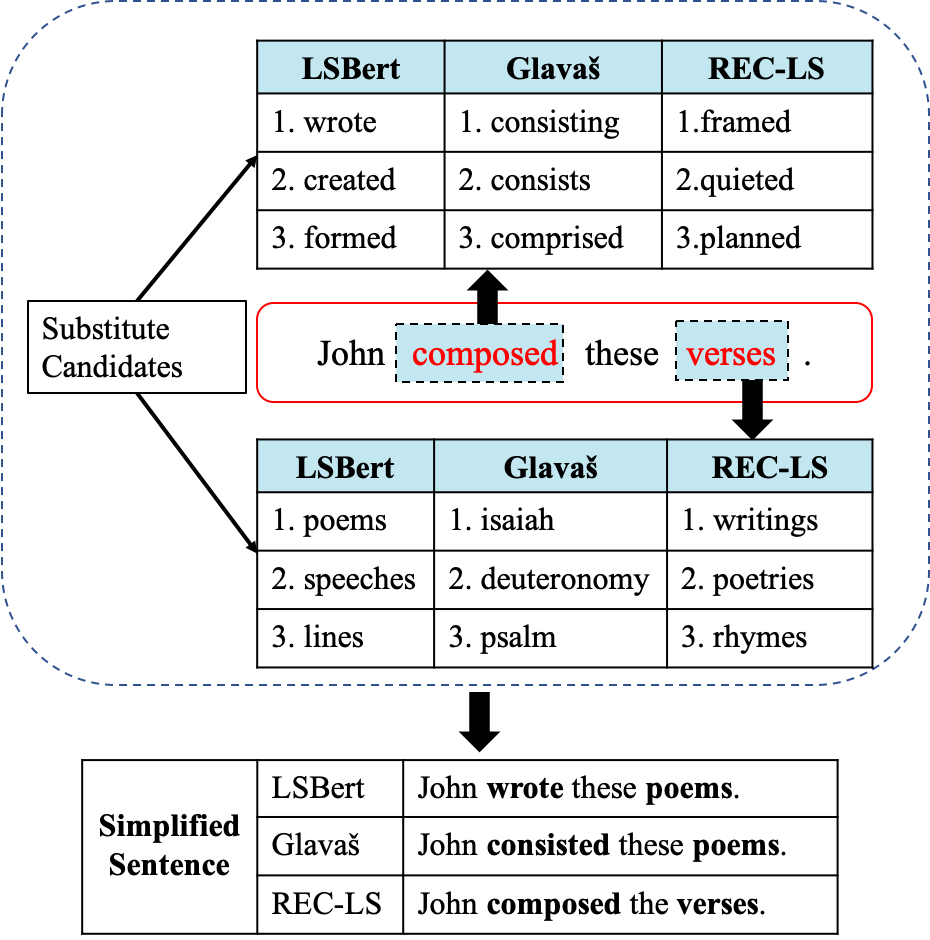}
  \caption{Comparison of substitute candidates of complex words. Given one sentence "John composed these verses." and complex words 'composed' and 'verses', the top three simplification candidates for each complex word are generated by our method LSBert and the state-of-the-art two baselines (Glava{\v{s}}\protect\cite{glavavs2015simplifying} and REC-LS \protect\cite{gooding2019recursive}). The simplified sentences by the three LS methods are shown at the bottom. }
  \label{Fig1}
\end{figure}

Context plays a central role in fulfilling substitute generation. Here, we give a simple example shown in Figure 1. For complex words 'composed' and 'verses' in the sentence "John composed these verses.", the top three substitute candidates of the two complex words generated by the state-of-the-art LS systems \cite{glavavs2015simplifying,gooding2019recursive} are only related with the complex words itself regardless of the context. For example, the candidates "consisting, consists, comprised" is generated by Glava{\v{s}}\cite{glavavs2015simplifying} for the complex word "composed", and the candidates "framed, quieted, planned" is produced by REC-LS \cite{gooding2019recursive}.

In contrast to the existing LS methods that only considered the context in the last step (substitute ranking), we present a novel LS framework LSBert, which takes the context into account in all three steps of LS. As word complexity depends on context, LSBert uses a novel approach to identify complex words using a sequence labeling method \cite{gooding2019complex} based on bi-directional long short-term memory units (BiLSTM). For producing suitable simplifications for the complex word, we exploit recent advances in pretrained unsupervised deep bidirectional representations Bert \cite{devlin2018bert} . More specifically, we mask the complex word $w$ of the original sentence $S$ as a new sentence $S'$, and concatenate the original sequence $S$ and $S'$ for feeding into the Bert to obtain the probability distribution of the vocabulary corresponding to the masked word. Then we choose the top probability words as substitute candidates. 

For ranking the substitutions, we adopt five high-quality features including word frequency and word similarity, Bert’s prediction order, Bert-based language model, and the paraphrase database PPDB, to ensure grammaticality and meaning equivalence to the original sentence in the output. LSBert simplifies one word at a time and is recursively applied to simplify the sentence by taking word complexity in context into account. As shown in Figure 1, the meaning of the original sentence using Glava{\v{s} is changed, and REC-LS does not make the right simplification. LSBert generates the appropriate substitutes and achieves its aim that replaces complex words with simpler alternatives.

This paper has the following two contributions:

(1) LSBert is a novel Bert-based method for LS, which can take full advantages of Bert to generate and rank substitute candidates. To our best knowledge, this is the first attempt to apply pretrained transformer language models for LS. In contrast to existing methods without considering the context in complex word identification and substitute generations, LSBert is easier to hold cohesion and coherence of a sentence, since LSBert takes the context into count for each step of LS

(2) LSBert is a simple, effective and complete LS method. 1)Simple: many steps used in existing LS systems have been eliminated from our method, e.g., morphological transformation. 2) Effective: it obtains new state-of-the-art results on three benchmarks. 3) Complete: LSBert recursively simplifies all complex words in a sentence without requiring additional steps.  

To facilitate reproducibility, the code of LSBert framework is available at https://github.com/BERT-LS.

The rest of this paper is organized as follows. In Section 2, we introduce the related work of text simplification. Section 3 describes the framework LSBert. In Section 4, we describe the experimental setup and evaluate the proposed method LSBert. Finally, we draw our conclusions in Section 5.

\section{Related Work}

Textual simplification (TS) is the process of simplifying the content of the original text as much as possible, while retaining the meaning and grammaticality so that it can be more easily read and understood by a wider audience. Textual simplification focuses on simplifying the vocabulary and syntax of the text. Early systems of TS often used standard statistical machine translation approaches to learn the simplification of a complex sentence into a simplified sentence \cite{coster2011simple}. Recently, TS methods adopted encoder-decoder model to simplify the text based on parallel corpora \cite{wang2016experimental,nisioi2017exploring,dong2019editnts}. All of the above work belong to the supervised TS systems, whose performance strongly relies on the availability of large amounts of parallel sentences. Two public parallel benchmarks WikiSmall \cite{zhu2010monolingual} and WikiLarge ~\cite{zhang2017sentence} contain a large proportion of: inaccurate simplifications (not aligned or only partially aligned) ; inadequate simplifications (not much simpler) ~\cite{xu2015problems,vstajner2015deeper}. These problems is mainly because designing a good alignment algorithm for extracting parallel sentences from EW and SEW is very difficult~\cite{hwang2015aligning}. Therefore, a number of approaches focusing on the generation and assessment of lexical simplification were proposed.

Lexical simplification (LS) only focuses to simplify complex words of one sentence. LS needs to identify complex words and find the best candidate substitution for these complex words \cite{shardlow2014survey,paetzold2017survey}. The best substitution needs to be more simplistic while preserving the sentence grammatically and keeping its meaning as much as possible, which is a very challenging task. The popular lexical simplification approaches were rule-based, in which each rule contains a complex word and its simple synonyms \cite{Lesk:1986:ASD:318723.318728,pavlick2016simple,maddela-xu-2018-word}. Rule-based systems usually identified synonyms from WordNet or other linguistic databases for a predefined set of complex words and selected the "simplest" from these synonyms based on the frequency of word or length of word \cite{devlin1998the,De_belder}. However, there is a major limitation for the rule-based systems that it is impossible to give all possible simplification rules for each word.

As complex and simplified parallel corpora are available, LS systems tried to extract rules from parallel corpora \cite{biran2011putting,yatskar2010sake,horn2014learning}. Yatskar et al. (2010) identified lexical simplifications from the edit history of simple English Wikipedia (SEW). They utilized a probabilistic method to recognize simplification edits distinguishing from other types of content changes. Biran et al. (2011) considered every pair of distinct word in the English Wikipedia (EW) and SEW to be a possible simplification pair, and filtered part of them based on morphological variants and WordNet. Horn et al. (2014) also generated the candidate rules from the EW and SEW, and adopted a context-aware binary classifier to decide whether a candidate rule should be adopted or not in a certain context. The main limitation of the type of methods relies heavily on parallel corpora. 

To entirely avoid the requirement of lexical resources or parallel corpora, LS systems based on word embeddings were proposed \cite{glavavs2015simplifying}. They extracted the top 10 words as candidate substitutions whose vectors are closer in terms of cosine similarity with the complex word. Instead of a traditional word embedding model, Paetzold and Specia (2016) adopted context-aware word embeddings trained on a large dataset where each word is annotated with the POS tag. Afterward, they further extracted candidates for the complex word by combining word embeddings with WordNet and parallel corpora \cite{paetzold2017lexical}. REC-LS \cite{gooding2019recursive} attempted to generate substitutes from multiple sources, e.g, WordNet, Big Huge Thesaurus \footnote{https://words.bighugelabs.com} and word embeddings.

After examining existing LS methods ranging from rules-based to embedding-based, the major challenge is that they generated simplification candidates for the complex word regardless of the context of the complex word, which will inevitably produce a large number of spurious candidates that confuse the systems employed in the subsequent steps. 

In this paper, we will first present a LS approach LSBert that requires only a sufficiently large corpus of raw text without any manual efforts. Pre-training language models \cite{devlin2018bert,lee2019biobert,lample2019cross} have attracted wide attention and has shown to be effective for improving many downstream natural language processing tasks. Our method exploits recent advances in Bert to generate suitable simplifications for complex words. Our method generates the candidates of the complex word by considering the whole sentence that is easier to hold cohesion and coherence of a sentence. In this case, many steps used in existing LS methods have been eliminated from our method, e.g., morphological transformation. The previous version LSBert was published in artificial intelligence conference (AAAI) \cite{qiang2020AAAI}, which only focused on substitute generations given the sentence and its complex word using Bert. In this paper, we propose an LS framework including complex word identification, substitute generations, substitute ranking. The framework can simplify one sentence recursively. One recent work for LS based Bert \cite{zhou2019bert} was almost simultaneously proposed with our previous version, which also only focused on substitute generations.

\section{Lexical Simplification Framework}

In this section, we outline each step of our lexical simplification framework LSBert as presented in Figure \ref{Fig2}, which includes the following three steps: complex word identification, substitute generation, filtering and substitute ranking. LSBert simlifies one complex word at a time, and is recursively applied to simplify the sentence. We will give the details of each step below.

\begin{figure}
  \centering
  \includegraphics[width=90mm]{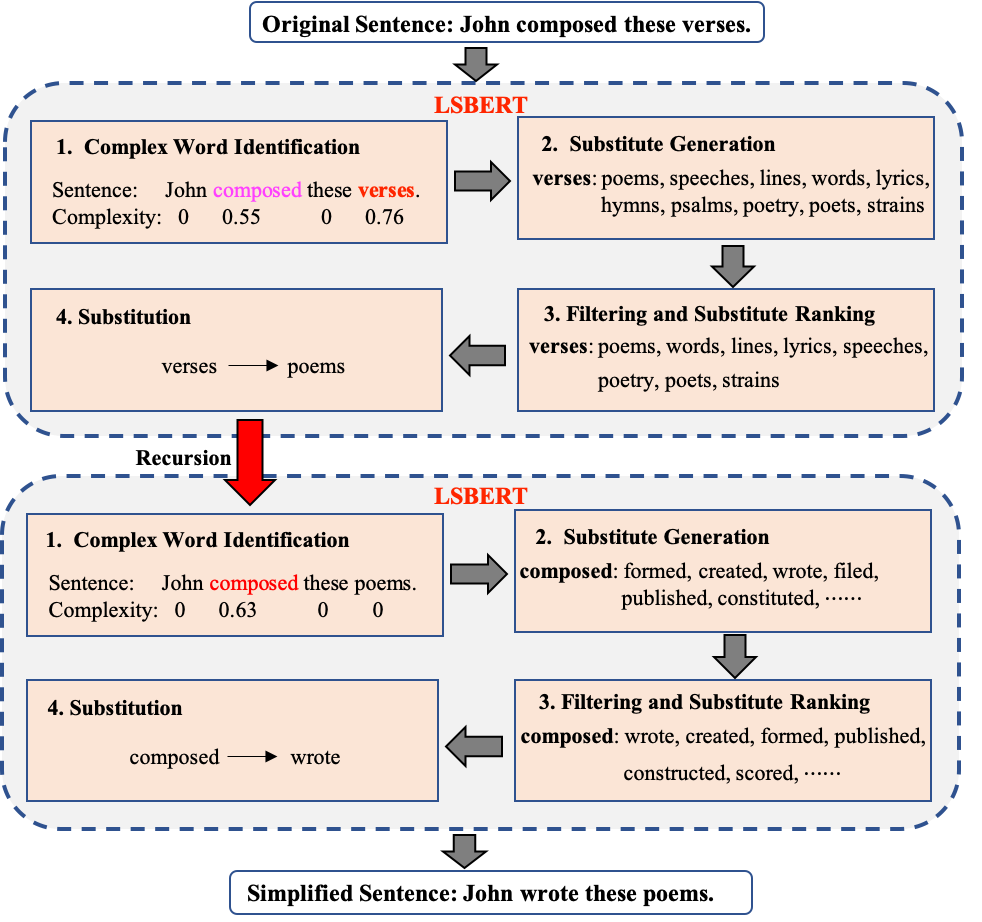}
  \caption{Overview of the lexical simplification framework LSBert.}
  \label{Fig2}
\end{figure}

\subsection{Complex Word Identification (CWI)}

Identifying complex words from one sentence has been studied for years, whose goal is to select the words in a given sentence which should be simplified \cite{shardlow2013comparison,yimam2018report}. 

CWI was framed as a sequence labeling task \cite{gooding2019complex} and an approach SEQ based on bi-directional long short-term memory units (BiLSTM) is trained to predict the binary complexity of words as annotated in the dataset of \cite{yimam2017cwig3g2}. In contrast to the other CWI models, the SEQ model has the following two advantages: takes word context into account and helps avoid the necessity of extensive feature engineering, because SEQ only relies on word embeddings as the only input information. 

The SEQ approach labels each word with a lexical complexity score ($p$) which represents the likelihood of each word belonging to the complex class. Giving a predefined threshold $p$, if the lexical complexity of one word is greater than the threshold, it will be treated as a complex word. For example, the example "John composed$_{0.55}$ these verses$_{0.76}$" is showed in Figure \ref{Fig2}. If the complexity threshold is set to 0.5, the two words "composed" and "verses" will be the complex words to be simplified. 

LSBert starts with the word "verses" with the highest $p$ value above the predefined threshold to simplify. After completing the simplification process, we will recalculate the complexity of each word in the sentence, excluding words that have been simplified. In addition, we exclude the simplification of entity words by performing named entity identification on the sentence.

\subsection{Substitute Generation (SG)}

Given a sentence $S$ and the complex word $w$, the aim of substitution generation (SG) is to produce the substitute candidates for the complex word $w$. LSBert produces the substitute candidates for the complex word using pretrained language model Bert. we briefly summarize the Bert model, and then describe how we extend it to do lexical simplification.

Bert \cite{devlin2018bert} is a self-supervised method for pretrained a deep transformer encoder, which is optimized by two training objectives: masked language modeling (MLM) and next sentence prediction (NSP). Unlike a traditional language modeling objective of predicting the next word in a sequence given the history, MLM predicts missing tokens in a sequence given its left and right context. Bert accomplishes NSP task by prepending every sentence with a special classification token, [CLS], and by combining sentences with a special separator token, [SEP]. The final hidden state corresponding to the [CLS] token is used as the total sequence representation from which we predict a label for classification tasks, or which may otherwise be overlooked.

\begin{figure*}
  \centering
  \includegraphics[width=170mm]{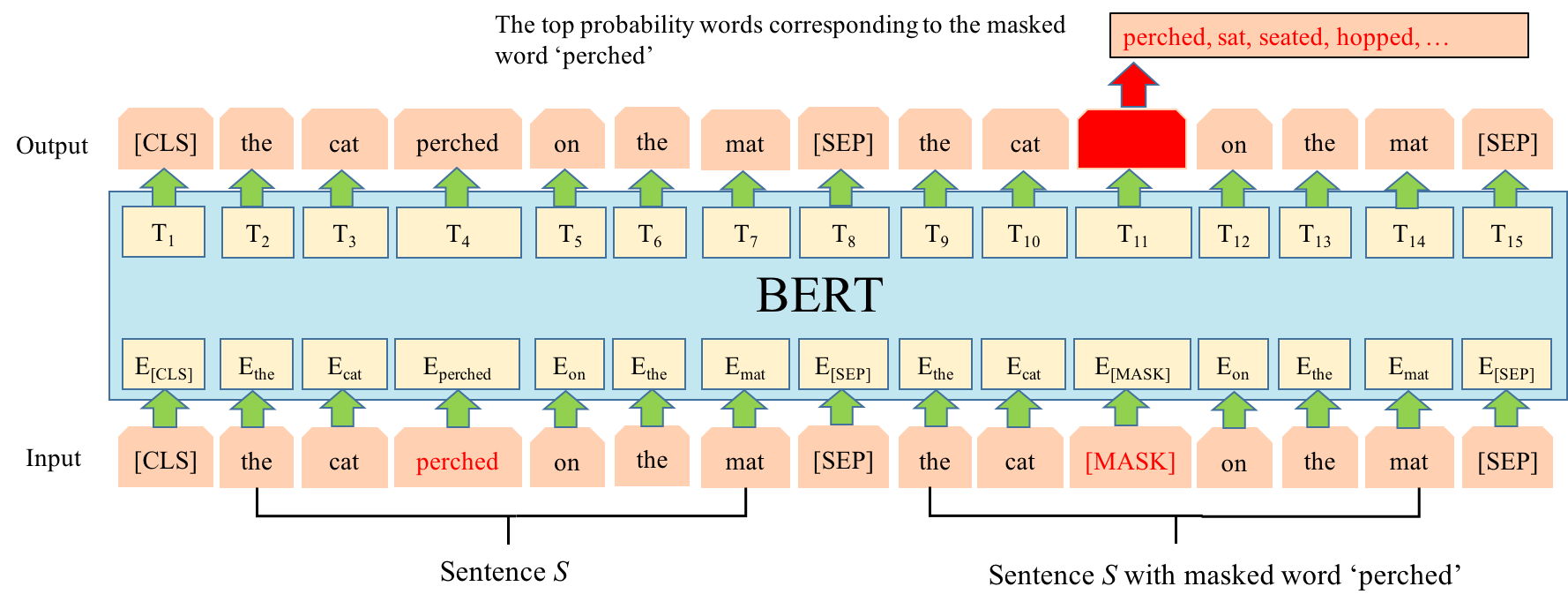},
  \caption{Substitution generation of LSBert for the target complex word prediction, or $cloze$ task. The input text is "the cat perched on the mat" with complex word "perched". [MASK], [CLS] and [SEP] are thress special symbols in Bert, where [MASK] is used to mask the word, [CLS] is added in front of each input instance and [SEP] is a special separator token. }
  \label{MLM}
\end{figure*} 

Due to the fundamental nature of MLM, we mask the complex word $w$ of the sentence $S$ and get the probability distribution of the vocabulary $p(\cdot|S\backslash \{w\})$ corresponding to the masked word $w$. Therefore, we can try to use MLM for substitute generation.

For the complex word $w$ in a sentence $S$, we mask the word $w$ of $S$ using special symbol "[MASK]" as a new sequence $S'$. If we directly feed $S'$ into MLM, the probability of the vocabulary $p(\cdot|S'\backslash \{t_i\})$ corresponding to the complex word $w$ only considers the context regardless of the influence of the complex word $w$. Considering that Bert is adept at dealing with sentence pairs due to the NSP task adopted by Bert. We concatenate the original sequence $S$ and $S'$ as a sentence pair, and feed the sentence pair $(S,S')$ into the Bert to obtain the probability distribution of the vocabulary $p(\cdot|S,S'\backslash \{w\})$ corresponding to the mask word. Thus, the higher probability words in $p(\cdot|S,S'\backslash \{w\})$ corresponding to the mask word not only consider the complex word itself, but also fit the context of the complex word. 

Finally, we select the top 10 words from $p(\cdot|S,S'\backslash \{w\})$ as substitution candidates, excluding the morphological derivations of $w$. In addition, considering that the contextual information of the complex word is used twice, we randomly mask a certain percentage of words in $S$ excluding $w$ for appropriately reducing the impact of contextual information. 

See Figure \ref{MLM} for an illustration. Suppose that there is a sentence "the cat perched on the mat" and the complex word "perched", we get the top three substitute candidates "sat, seated, hopped".  We can see that the three candidates not only have a strong correlation with the complex word, but also hold the cohesion and coherence properties of the sentence. If we adopt the existing state-of-the-art methods \cite{glavavs2015simplifying} and \cite{gooding2019recursive}, the top three substitution words are "atop, overlooking, precariously" and "put, lighted, lay", respectively. Very obviously, our method generates better substitute candidates for the complex word.

\subsection{Filtering and Substitute Ranking (SR)}

Giving substitute candidates $C=\{c_1,c_2,...,c_n\}$, the substitution ranking of the lexical simplification framework is to decide which one of the candidate substitutions that fits the context of complex word is the simplest \cite{paetzold2017survey}, where $n$ is the number of substitute candidates. First, threshold-based filtering is performed by LSBert, which is used to remove some complex substitutes. Substitutes are removed from consideration if their Zipf values below 3 using Frequency features. Then, LSBert computes various rankings according to their scores for each of the features. After obtaining all rankings for each feature, LSBert scores each candidate by averaging all its rankings. Finally, we choose the candidate with the highest ranking as the best substitute.

Previous work for this step is based on the following features: word frequency, contextual simplicity and N\-gram language modeling, etc. In contrast to previous work, in addition to the word frequency and word similarity commonly used in other LS methods, LSBert considers three additional high-quality features: two features about Bert and one feature about PPDB (A Paraphrase Database for Simplification). 

\textbf{Bert prediction order.} On this step of substitute generation, we obtain the probability distribution of the vocabulary corresponding to the mask word. Because LSBert already incorporates the context information on the step of substitution generation, the word order of Bert prediction is a crucial feature which includes the information of both the context and the complex word itself. The higher the probability, the more relevant the candidate for the original sentence. 

\textbf{Language model feature.} A substitution candidate should fit into the sequence of words preceding and following the original word. We cannot directly compute the probability of a sentence or sequence of words using Bert like traditional n-gram language models. Let $W=w_{-m},...,w_{-1},w,w_1,...,w_m$ be the context of the original word $w$. We adopt a new strategy to compute the likelihood of $W$. We first replace the original word $w$ with the substitution candidate. We then mask one word of $W$ from front to back and feed into Bert to compute the cross-entropy loss of the mask word. Finally, we rank all substitute candidates based on the average loss of $W$. The lower the loss, the substitute candidate is a good substitution for the original word. We use as context a symmetric window of size five around the complex word.

\textbf{Semantic similarity.} The similarity between the complex word and the substitution candidate is widely used as a feature for SR. In general, word embedding models are used to obtain the vector representation and the cosine similarity metric is chosen to compute the similarity. Here, we choose the pretrained fastText model \footnote{https://dl.fbaipublicfiles.com/fasttext/vectors-english/crawl-300d-2M-subword.zip} as word embedding modeling. The higher the similarity value, the higher the ranking.

\textbf{Frequency feature.} Frequency-based candidate ranking strategies are one of the most popular choices by lexical simplification and quite effective. In general, the more frequency a word is used, the most familiar it is to readers. We adopt the Zipf scale created from the SUBTLEX lists \cite{Brysbaert2009Moving}, because some experiments \cite{paetzold2017survey} revealed that word frequencies from this corpus correlate with human judgments on simplicity than many other more widely used corpora, such as Wikipedia. SUBTLEX \footnote{http://subtlexus.lexique.org} is composed of over six million sentences extracted from subtitles of assorted movies. The Zipf frequency of a word is the base-10 logarithm of the number of times it appears per billion words. 

\textbf{PPDB feature.} Some LS methods generated substitute candidates from PPDB or its subset SimplePPDB \cite{pavlick2016simple,Kriz2018Simplification}. PPDB is a collection of more than 100 million English paraphrase pairs \cite{Ganitkevitch2013}. These pairs were extracted using a bilingual pivoting technique, which assumes that two English phrases that translate to the same foreign phrase have the same meaning. Since LSBert has a better substitution generation than PPDB and SimplePPDB, they cannot help improve the performance of substitution generation. Considering PPDB owns useful information about paraphrase, we try to use PPDB as a feature to rank the candidate substitutions. We adopt a simple strategy for PPDB to rank the candidates. For each candidate $c_i$ in $C$ of $w$, the ranking of $c_i$ is 1 if the pair $(w,c_i)$ exists in PPDB. Otherwise, the ranking number of $c_i$ is $n/3$.

\subsection{LSBert Algorithm}

Following CWI, substitute generation, filtering and substitute ranking steps, the overall simplification algorithm LSBert is shown in Algorithm 1 and Algorithm 2. Given the sentence $S$ and complexity threshold $t$, we first identify named entity using entity identification system \footnote{https://spacy.io/}. We add entities into $ignore\_list$ which means these words do not need to be simplified. 

 In LSBert, we identify all complex words in sentence $s$ using CWI step excluding $ignore\_list$ (line 1). If the number of complex words in the sentence $s$ is larger than 0 (line 2), LSBert will try to simplify the top complex word $w$ (line 3). LSBert calls substitute generation (line 4) and substitute ranking (line 5) in turn. LSBert chooses the top substitute (line 6). One important thing to notice is whether LSBert performs the simplification only if the top candidate $top$ has a higher frequency (Frequency feature) or lower loss (Language model feature) than the original word (line 7). When LSBert performs the simplification, it will replace $w$ into $top$ (line 8) and add the word $top$ into $ignore\_list$ (line 9). After completing the simplification of one word, we will iteratively call LSBert (line 10 and line 12). If the number of complex words in $S$ equals to 0, we will stop calling LSBert (line 15).

\begin{algorithm}[tb]
\caption{ Lexical simplification framework}
\label{alg:algorithm}
\begin{algorithmic}[1] 
\STATE $S$ $\leftarrow$ Input Sentence
\STATE $t$ $\leftarrow$ Complexity threshold
\STATE $ignore\_list$ $\leftarrow$ Named$\_$Entity$\_$Identification($S$)
\STATE LSBert($S$,$t$,$ignore\_list$)
\end{algorithmic}
\end{algorithm}

\begin{algorithm}[tb]
\caption{ LSBert ($S$,$t$,$ignore\_list$)}
\label{alg:algorithm}
\begin{algorithmic}[1] 
\STATE $complex\_words$ $\leftarrow$ CWI($S$,$t$)-$ignore\_list$
\IF {number($complex\_words$)$>$0}
\STATE $w$ $\leftarrow$ head($complex\_words$)
\STATE $subs$ $\leftarrow$ Substitution$\_$Generation($S$,$w$)
\STATE $subs$ $\leftarrow$ Substitute$\_$Ranking($subs$)
\STATE $top$ $\leftarrow$ head($subs$)
\IF{$fre(top$)$>$$fre(w)$ or $loss(top$)$<$$loss(w)$}
\STATE Replace($S$,$w$,$top$)
\STATE $ignore\_list$.add($w$)
\STATE LSBert($S$,$t$,$ignore\_list$)
\ELSE 
\STATE LSBert($S$,$t$,$ignore\_list$)
\ENDIF
\ELSE
\STATE return $S$
\ENDIF
\end{algorithmic}
\end{algorithm}

\section{Experiments}

We design experiments to answer the following questions:

\textbf{Q1. The effectiveness of substitute candidates and ranking:} Does the simplification candidate generation of LSBert outperforms the substitution generation of the state-of-the-art competitors?

\textbf{Q2. The effectiveness of the LS system: } Do the  of LSBert outperforms the full pipeline of the state-of-the-art competitors?

\textbf{Q3. The factors of affecting the LSBert: } Experiments on different parameters and models verify the impact on the LSBert system.

\textbf{Q4. The qualitative study of the LSBert: } We do more experiments to analyze the advantages and the disadvantages of LSBert.

\textbf{Dataset}. We choose the following datasets to evaluate our framework LSBert from lexical simplification datasets and text simplification dataset.

(1) We use three widely used lexical simplification datasets (LexMTurk\footnote{http://www.cs.pomona.edu/~dkauchak/simplification/lex.mturk.14} \cite{horn2014learning},  BenchLS \footnote{http://ghpaetzold.github.io/data/BenchLS.zip} \cite{paetzold2016unsupervised}, NNSeval \footnote{http://ghpaetzold.github.io/data/NNSeval.zip} \cite{paetzold2017survey}) to do experiments. The details of the three datasets are illustrated in this paper \cite{paetzold2017survey}. Notice that, because these datasets already offer the target words regarded complex by human annotators, we do not address complex word identification task in our evaluations using the three datasets. These datasets contain instances composed of a sentence, a target complex word, and a set of suitable substitutions provided and ranked by humans with respect to their simplicity.

(2) We use one widely used text simplification dataset (WikiLarge) to do experiments \cite{zhang2017sentence}. The training/development/test set in WikiLarge have 296,402/2000/359 sentence pairs, respectively. WikiLarge is a set of automatically aligned complex-simple sentence pairs from English Wikipedia (EW) and Simple English Wikipedia (SEW). Its validation and test sets are taken from Turkcorpus, where each original sentence has 8 human simplifications created by Amazon Mechanical Turk workers.

\begin{table*}
\centering
\begin{tabular}{l|ccc|ccc|ccc|}
\hline
& \multicolumn{3}{|c|}{LexMTurk}  & \multicolumn{3}{|c|}{BenchLS}  & \multicolumn{3}{|c|}{NNSeval} \\

 &   PRE & RE & F1 & PRE & RE & F1 & PRE & RE & F1\\
\hline
Yamamoto & 0.056 &  0.079 & 0.065  & 0.032 & 0.087 & 0.047 & 0.026 & 0.061 &  0.037 \\
Devlin & 0.164 &  0.092 & 0.118 & 0.133 & 0.153 & 0.143 & 0.092 & 0.093 & 0.092  \\
Biran &  0.153 & 0.098  & 0.119  & 0.130 & 0.144 & 0.136 & 0.084 & 0.079 & 0.081  \\
Horn &  0.153 & 0.134  & 0.143 & 0.235 & 0.131 & 0.168 & 0.134 & 0.088 & 0.106  \\
Glava{\v{s}} &  0.151 &  0.122 &  0.135 & 0.142 & 0.191 & 0.163 & 0.105 & 0.141 & 0.121 \\ 
Paetzold-CA &  0.177 & 0.140  & 0.156  & 0.180 & 0.252 & 0.210 & 0.118 & 0.161 & 0.136 \\
Paetzold-NE &  \textbf{0.310} & 0.142  &  0.195 & \textbf{0.270} & 0.209 & 0.236 & 0.186 & 0.136 & 0.157 \\
REC-LS & 0.151 & 0.154 & 0.152 & 0.129 & 0.246 & 0.170 & 0.103 & 0.155 & 0.124 \\
\hline
Bert-mask & 0.254 & 0.197&0.222 & 0.176 & 0.239 & 0.203 & 0.138 & 0.185 & 0.158 \\
Bert & 0.256 & 0.199 &  0.224 & 0.210 &0.285 & 0.242& 0.154 & 0.205 & 0.176 \\
Bert-dropout & 0.255 & 0.198 & 0.223 & 0.204 & 0.277 & 0.235 & 0.153 & 0.204 & 0.175 \\
\hline
LSBert$_{pre}$ &  0.287 & 0.223 & 0.251 & 0.231 & 0.314 & 0.267 &  0.185 & 0.246 & 0.211 \\
LSBert &  0.306 & \textbf{0.238} & \textbf{0.268} & 0.244 & \textbf{0.331} & \textbf{0.281} &  \textbf{0.194} & \textbf{0.260} & \textbf{0.222} \\
\hline
\end{tabular}
\caption{Evaluation results of substitute generation on three datasets. }
\label{SGResults}
\end{table*}

\subsection{Experiment Setup}

We choose the following baselines to comparison:.

(1) Linguistic databases. \textbf{Devlin} \cite{devlin1998the} extracts synonyms of complex words from WordNet. \textbf{Yamamoto} \cite{kajiwara2013selecting} is proposed for Japanese based on dictionary definitions to extract substitute candidates. Here, Yamamoto is adapted for English by using the Merriam Dictionary to extract definitions of complex words. 

(2) Parallel corpus. \textbf{Biran} \cite{biran2011putting} and \textbf{Horn} \cite{horn2014learning} perform substitute generation (SG) through parallel corpora EW and SEW.

(3) Paraphrase database. \textbf{SimplePPDB} \cite{pavlick2016simple} performs SG with a filtered paraphrase database (PPDB).

(4) Word embeddings. \textbf{Glava{\v{s}}} \cite{glavavs2015simplifying} performs SG with typical word embeddings. \textbf{Paetzold-CA} \cite{paetzold2016unsupervised} performs SG with context-aware word embeddigns.

(5) Multipe source. \textbf{Paetzold\-NE} \cite{paetzold2017lexical} performs SG with parallel corpora and context-aware word embeddigns. \textbf{REC-LS} \cite{gooding2019recursive} performs SG with typical word embeedings and linguistic databases.

(6) Methods based on Bert. Here, we give multiple strategies to perform SG using Bert. \textbf{Bert-mask}: we directly mask the complex word of the sentence and feed it into Bert. \textbf{Bert}: we directly feed the sentence into Bert to generate substitute generates. \textbf{Bert-dropout} \cite{zhou2019bert} applied dropout to the complex word's embeddings for partially masking the word. These Bert-based baselines are based on the single sentence that uses to feed into Bert.  

(7) Our proposed methods. \textbf{LSBert$_{pre}$} is our previous version. \textbf{LSBert} is the proposed method in this paper. Our two methods LSBert$_{pre}$ and LSBert feed two sentences for Bert. 

The experimental results of Devlin, Yamamoto, Biran, Horn, and SimplePPDB, Glava{\v{s}, Paetzold-CA, and Paetzold-NE are from these two papers \cite{paetzold2017survey,paetzold2017lexical}. For REC-LS method, we use the code proposed by the authors. \textbf{Bert-dropout} was re-implemented based on the original paper. In all experiments for methods based on Bert, we use BERT-Large, Uncased (Whole Word Masking) pre-trained on BooksCorpus and English Wikipedia \footnote{https://github.com/google-research/bert}.

\subsection{Quantitative Evaluation}

\textbf{ (1) Evaluation of Substitute Candidates}

The following three widely used metrics are used for evaluation \cite{paetzold2015lexenstein,paetzold2016unsupervised,paetzold2017survey}. 

\textbf{Precision (PRE)}: The proportion of generated candidates that are in the gold standard. 

\textbf{Recall (RE)}: The proportion of gold-standard substitutions that are included in the generated substitutions. 

\textbf{F1}: The harmonic mean between Precision and Recall.

The results are shown in Table \ref{SGResults}. As can be seen, our model LSBert obtains the highest Recall and F1 scores on three datasets, largely outperforming the previous best baseline Paetzold-NE, increasing 37.4\%, 19.1\% and 41.4\% using F1 metric. The baseline Paetzold-NE by combining the Newsela parallel corpus and context-aware word embeddings obtains better results on PRE than LSBert, because it uses a different calculation method. If one candidate exists in the gold standard, different morphological derivations of the candidate in substitute candidates are all counted into the PRE metric. Because of considering the context, the substitute candidates of Bert based methods are normally different words.

We note that the Bert based model is not only able to outperform other systems on all datasets using F1, but it also
has two clear practical advantages: (1) the only input information it uses at run time is Bert without requiring linguistic database and comparable corpus, (2) the substitute candidates using Bert do not require morphological transformation.

For these baselines based on a single sentence (Bert-mask, Bert and Bert-dropout), the gap between them is very small. Compared with Bert based on a single sentence, our method LSBert$_{pre}$ and LSBert have better results, which verify that our strategy based on sentence pairs fits for lexical simplification. In conclusion, the results clearly show that LSBert provides a good balance precision and recall using only Bert.

\textbf{ (2) Evaluation of SG and SR}

\begin{table}
\centering
\begin{tabular}{l|cc|cc|cc}
\hline
& \multicolumn{2}{|c|}{LexMTurk}  & \multicolumn{2}{|c|}{BenchLS}  & \multicolumn{2}{|c}{NNSeval} \\
 &  PRE & ACC &  PRE & ACC & PRE & ACC   \\
\hline
Yamamoto & 0.066& 0.066  &  0.044 & 0.041 & 0.444 & 0.025 \\
Biran &0.714 & 0.034  &  0.124  & 0.123 & 0.121 & 0.121 \\
Devlin & 0.368 &  0.366 &  0.309  & 0.307 &  0.335 & 0.117  \\
Paetzold\-CA & 0.578 & 0.396  &  0.423 & 0.423 & 0.297 & 0.297    \\ 
Horn & 0.761 & 0.663  & 0.546  & 0.341 & 0.364 &  0.172 \\
Glava{\v{s}} & 0.710 & 0.682  & 0.480  & 0.252 & 0.456 & 0.197   \\
Paetzold\-NE & 0.676 &  0.676 & 0.642 & 0.434 & 0.544  & 0.335  \\
REC-LS  & 0.786 & 0.256 & \textbf{0.734} & 0.335 & \textbf{0.665} & 0.218 \\
\hline
LSBert$_{pre}$  & 0.770  & 0.770  & 0.604 & 0.604 & 0.420 & 0.420\\
LSBert  & \textbf{0.864} & \textbf{0.792} & 0.697 & \textbf{0.616} & 0.526 & \textbf{0.436}\\
\hline
\end{tabular}
\caption{The evaluation results using Precision (PRE) and Accuracy (ACC) on three datasets.}
\label{pipeline}
\end{table}

In this section, we evaluate the performance of various LS systems that combines SG and SR. We adopt the following two well-known metrics used by these work \cite{horn2014learning,paetzold2017survey}. 

\textbf{Precision (PRE)}: The proportion with which the replacement of the original word is either the original word itself or is in the gold standard. 

\textbf{Accuracy (ACC)}: The proportion with which the replacement of the original word is not the original word and is in the gold standard.  

It can be seen from these two metrices that if no simplification is carried out, the PRE value is 1 and the ACC value is 0. If all complex words are replaced by the substitutions, the PRE and ACC vaule have the same value.

The results are shown in Table \ref{pipeline}. We can see that our method LSBert attains the highest Accuracy on three datasets, which has an average increase of 29.8\% over the former state-of-the-art baseline (Paetzold-NE). It suggests that LSBert is the most proficient in promoting simplicity. Paetzold-NE obtains higher than LSBert on Precision on NNSeval, which also means that many complex words are replaced by the original word itself, due to the shortage of simplification rules in parallel corpora. REC-LS obtains the best PRE and poor ACC, because it prefers the original word as the substitute word. 

 In conclusion, although LSBert only uses raw text for pre-trained Bert without using any resources, LSBert remains the best lexical simplification method. The results are in accordance with the conclusions of the Substitute Generation.

\textbf{ (3)Evaluation of LS system for sentence simplification}

Lexical simplification evaluation needs to be provided with the sentence and the specified complex word. Here, we try to simplify one sentence instead of one word, and choose a sentence simplification dataset (WikiLarge) for evaluation. 

Since most LS methods only focused on one or two steps (SR or SR) of LS, they cannot directly simplify one setence. Here, we choose two complete LS systems (Glava{\v{s}} \cite{glavavs2015simplifying} and REC-LS \cite{gooding2019recursive} ) to comparison. In additional, we choose four state-of-the-art text simplification (TS) methods DRESS-LS \cite{zhang2017sentence}, EditNTS \cite{dong2019editnts}, PBMT \cite{Jipeng2019Unsupervised}, and Access \cite{martin2019controllable}. The first three TS methods except PBMT are sequence-to-sequence modelings and all need training data sets to learn. PBMT is an unsupervised text simplification system based on phrase-based machine translation system. For LS methods, they only use the testset to output the simplified sentences. For LSBert and Rec-LS, the complexity threshold of CWI is 0.5. For Glava{\v{s} method, it tries to simplify all content words (noun, verb, adjective, or adverb) of one sentence.

Following previous work, two widely used metrics (SARI and FRES) in text simplification are chosen in this paper ~\cite{woodsend2011learning,xu2016optimizing}. SARI \cite{zhang2017sentence} is a text-simplification metric by comparing the output against the simple and complex simplifications \footnote{ We used the implementation of SARI in \cite{xu2016optimizing}.}. Flesch reading ease score (FRES) measures the readability of the output \cite{kincaid1975}. A higher FRES represents simpler output.

\begin{table}
\centering
\begin{tabular}{|l|l|cc|}
\hline
 & Methods& SARI  & FRES  \\ \hline
 \multirow{4}*{TS methods}&DRESS-LS (2017) & 37.27 &  75.33 \\
 &EditNTS (2019)  & 38.22 & 73.81 \\
 &PBMT (2020) & 39.08 & 76.50 \\
 & Access (2019) & \textbf{41.87} & \textbf{81.55} \\
\hline

\multirow{3}*{LS methods} &Glava{\v{s}} & 30.70 & \textbf{81.82} \\
&REC-LS & 37.11 &  69.58 \\
&LSBert & \textbf{39.37} &  77.07 \\
\hline
\end{tabular}
\caption{ Comparison of text simplification methods on WikiLarge dataset.}
\label{TS}
\end{table}

Table \ref{TS} shows the results of all models on WikiLarge dataset. Our model LSBert obtains a SARI score of 39.37 and a FRES score of 77.07, even outperforming these three supervised TS systems (DRESS-LS, EditNTS and PBMT), which indicates that the model has indeed learned to simplify the complex sentences. Compared with LS methods Glava{\v{s} and REC-LS, LSBert achieves the best results. The two methods go to two different extremes, in which Glava{\v{s} simplifies almost all content words of one sentence and REC-LS prefers to save the original word. On the FRES metric, we can see that Glava{\v{s} outperforms LSBert, which is also because it simplifies almost all content words without caring for the equivalent meaning with the original sentence. Compared with Access, our model is highly competitive, because LSBert does not need a parallel dataset to learn and only focuses on simplifying the words. In conclusion, we can see that LSBert outperforms previous LS baselines, even some supervised TS baselines, which indicate that our method is effective at creating simpler output.

\subsection{Ablation Study of LSBert}

To further analyze the advantages and the disadvantages of LSBert, we do more experiments in this section.

\textbf{(1) Influence of Ranking Features}

\begin{table}
\centering
\begin{tabular}{|l|cc|cc|cc|cc|}
\hline
& \multicolumn{2}{|c|}{LexMTurk}  & \multicolumn{2}{|c|}{BenchLS}  & \multicolumn{2}{|c|}{NNSeval} \\ \hline
 &  PRE & ACC &  PRE & ACC & PRE & ACC   \\ \hline
LSBert & \textbf{0.864} & \textbf{0.792} & 0.697 & 0.616 & 0.526 & \textbf{0.436} \\
\hline
w/o Bert&        0.828 &  0.774  &  0.680  &  0.629 & 0.456 & 0.406  \\ 
w/o Language&    0.828 &  0.744  &  0.670  &  0.610 & 0.527 & 0.418  \\
w/o Similarity&  0.820 &  0.768  &  0.659  &  0.607 & 0.452 & 0.397  \\
w/o Frequency&   0.842 &  0.694  &  \textbf{0.713}  &  0.554 & \textbf{0.556} & 0.393 \\
w/o PPDB     &   0.852 &  0.784  &  0.698  &  \textbf{0.622} & 0.502 & 0.422  \\\hline
\end{tabular}
\caption{ Ablation study results of the ranking features.}
\label{rankingfeatures}
\end{table}

To determine the importance of each ranking feature, we make an ablation study by removing one feature in turn. The results are presented in Table \ref{rankingfeatures}. We can see that LSBert combining all five features achieves the best results, which means all features have a positive effect. LSBert removing frequency feature achieves better results on PRE metric, but it decreases the values of ACC. These features have different contributions for LSBert's performance, for example, PPDB feature brings the least impact on the performance of LSBert compared with the other features. In this paper, LSBert thinks all features are equally important, that may not be the best option. In the future, we can improve LSBert by combining these features using different weights. 

\textbf{ (2) Influence of Different Bert Modeling for Substitute Generation}

Pretrained Bert plays one vital role in LSBert. Bert has different versions based on the parameter scale and training strategy. Here, we attempt to investigate the influence of different Bert versions on the performance of LSBert. We choose the following three Bert models: 

(1) Bert-based, uncased (Base): 12-layer, 768-hidden, 12-heads, 110M parameters.

(2) Bert-large, uncased (Large): 24-layer, 1024-hidden, 16-heads, 340M parameters.

(3) Bert-large, uncased, Whole Word Masking (WWM): 24-layer, 1024-hidden, 16-heads, 340M parameters. The above two Bert models randomly select WordPiece tokens to mask. Whole Word Masking always masks all of the tokens corresponding to a word at once.

\begin{table}
\centering
\begin{tabular}{|l|l|ccc|cc|}
\hline
& & \multicolumn{3}{|c|}{\textbf{SG}}&\multicolumn{2}{|c|}{\textbf{SR}} \\
\hline
  & & PRE & RE &  F1  &  PRE & ACC  \\ \hline

  \multirow{2}*{LexMTurk}  & Base & 0.317 & 0.246 & 0.277 & 0.744 & 0.704 \\
   & Large& \textbf{0.333} & \textbf{0.259}  & \textbf{0.291} & 0.792 & 0.750 \\
   & WWM & 0.306 & 0.238 & 0.268 & \textbf{0.864} & \textbf{0.792} \\
\hline
  \multirow{2}*{BenchLS} & Base  & 0.233 & 0.317 & 0.269 & 0.586 & 0.537  \\
  & Large & \textbf{0.252} & \textbf{0.342}  &  \textbf{0.290} & 0.636 &  0.589 \\
  & WWM &  0.244 & 0.331 & 0.281 &  \textbf{0.697} & \textbf{0.616}  \\ \hline

  \multirow{2}*{NNSeval} &  Base & 0.172 & 0.230 &  0.197   & 0.393 &    0.347     \\
  &  Large & 0.185 & 0.247 & 0.211 & 0.402  &   0.360         \\
  & WWM & \textbf{0.194} & \textbf{0.260} & \textbf{0.222}  & \textbf{0.526} & \textbf{0.436} \\
\hline
\end{tabular}
\caption{Influence of different Bert models.}
\label{diff_bert}
\end{table}

Table \ref{diff_bert} shows the results of the experiments using different Bert models on three datasets. From Table \ref{diff_bert}, we can see that the WWM model obtains the highest accuracy and precision over the two other models. Besides, the Large model outperforms the Base model. It can be concluded that a better Bert model can help to improve the performance of LSBert system. If in the future a better Bert model is available, one can try to replace the Bert model in this paper to further improve the performance of LS system.

\textbf{(3) Influence of the Number of Substitute Candidates}

\begin{figure*} 
\begin{minipage}{0.3\linewidth}
\centerline{\includegraphics[width=6cm]{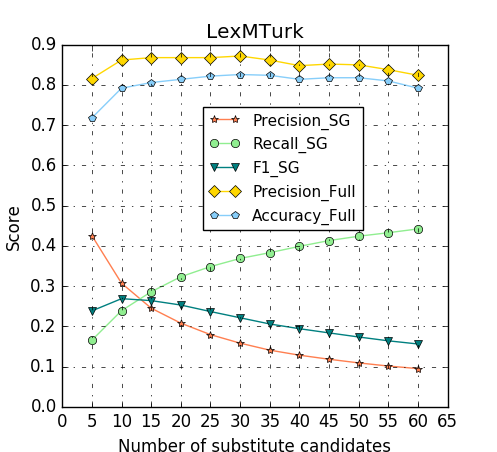}}
\end{minipage}
\hfill\begin{minipage}{.3\linewidth}
\centerline{\includegraphics[width=6cm]{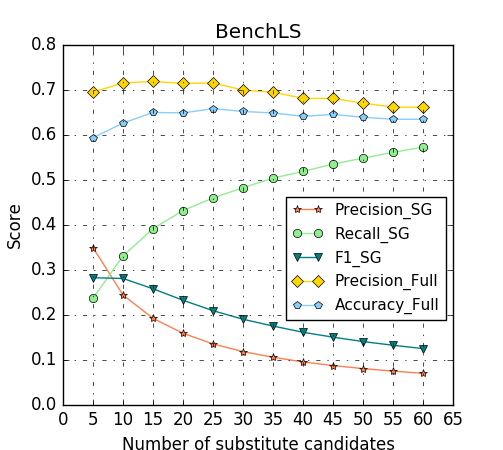}}
\end{minipage}
\hfill\begin{minipage}{0.3\linewidth}
\centerline{\includegraphics[width=6cm]{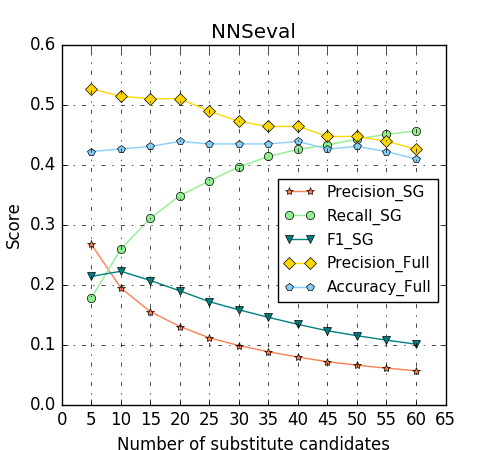}}
\end{minipage}
\caption{ Influence of number of substitute candidates.} \label{parameter}
\end{figure*}

In this part, we try to investigate the influence of the number of simplification candidates to the performance of LSBert. The number of candidates ranges from 5 to 60, respectively. Figure \ref{parameter} shows the performance of substitute candidates (Precision, Recall and F1), SR and SR (Precision, Accuracy) varying the number of candidates on three benchmarks. When increasing the number of candidates, the score of precision decreases and the score of recall increases. When increasing the number of candidates, the score of F1 first increases, and declines finally. The best performance of LSBert through the experiments is achieved by setting the number of candidates equals 10 for a good trade-off between precision and recall. The score of the accuracy and precison of the SG and SR (full) first increase and converge finally, which means that the SG and SR is less sensitive to the number of candidates. 

\begin{table*}
\centering
\begin{tabular}{l|l}
\hline
Sent1 &  Much of the water carried by these streams is \textbf{diverted} . \\

Labels & drawn away, redirected, changed, turned, moved, rerouted, led away, sent away, separated, switched, split, ...\\

LSBert & reclaimed, displaced, transferred, derived, pumped, routed, converted, recycled, discarded, drained \\
\hline
Sent2 & ... , every person born into the world is enslaved to the service of sin and , apart from the \textbf{efficacious} or prevenient grace of God, ...  \\

Labels & ever, present, showy, useful, effective, capable, strong, valuable, powerful, active, efficient, helpful, generous, power, kindness, effect, ... \\

LSBert & benevolent, exemplary, abundant, extraordinary, essential, inspired, ubiquitous, irresistible, exclusive, inclusive  \\
\hline
Sent3 &  The Amazon Basin is the part of South America drained by the Amazon River and its \textbf{tributaries} . \\

Labels &  streams, branches, riverlets, adjacent, smaller rivers, channels, rivers, brooks, ditches, children creeks, offshoots, creeks\\

LSBert &  basins, drains, derivatives, headwaters, components, subsidiaries, minions, rays, sources, forks \\
\hline
Sent4 &  He held several \textbf{senior} positions in the Royal Flying Corps during World War I, ... \\

Labels & high-level, older, upper, top, higher, high, superior, important, veteran, head, advance, top-level, advanced, leader, chief, principal, big  \\

LSBert & junior, significant, prestigious, leadership, civil, command, formal, prominent, subordinate, powerful \\
\hline
 
Sent5 & On 1 October 1983 the pilot project began operations as a \textbf{full-fledged} bank and was renamed the Grameen Bank to ...  \\

Labels &  real, developed, fully operating, legitimate, total, complete, full, qualified, whole, major, working, full-service, ... \\

LSBert &  commercial, development, national, community, central, formal, private, public, bangladeshi, chartered\\
\hline

Sent6 &  The principal greenhouse , in an art nouveau style with ... , \textbf{resembles} the mid-19th century Crystal Palace in London . \\

Labels & is similar to, looks like, looks-like, mimics, represents, matches, shows, mirrors, echos, look like, favors, appears like, ... \\

LSBert & recalls, suggests, approaches, echoes, references, parallels, appears, depicts, incorporates, follows \\
\hline

Sent7 & A perfectly elastic collision is defined as one in which there is no loss of \textbf{kinetic} energy in the collision . \\

Labels & active, moving, movement, motion, static, motive, innate, kinetic, real, strong, driving, motion related, motion-, living, powerful, ... \\

LSBert &  mechanical, rotational, dynamic, total, thermal, momentum, physical, the, potential, energetic \\
\hline

Sent8 &  None of your watched items were \textbf{edited} in the time period displayed . \\
Labels &  changed, looked at, refined, revise, finished, fixed, revised, revised, scanned, shortened  \\
LSBert & altered, incorporated, appropriate, modified, organized, filtered, included, blended, amended, enhanced \\
\hline
\end{tabular}
\caption{The examples of substitute candidates that do not contain one valid substitution provided by humans on LexMTurk. The complex word of each sentence are shown in bold.}
\label{sg_example}
\end{table*}

\subsection{Qualitative Study}

All of the above experiments are quantitative analyses of LSBert. Here, we also qualitatively evaluate our model from three aspects: substitute generation, substitute ranking and sentence simplification. 

\textbf{(1) The analysis of substitute generation results}

When the number of substitute candidates is set to 10, the proportion of LSBert that generates at least one valid substitute candidate is 98.6\% on Lexmturk dataset, namely, LSBert only produces no effective substitute word in only 8 sentences. When the number of generated candidates is 15, LSBert cannot generate any valid candidates on only 4 sentences. When the number of generated candidates is 30, only one sentence cannot be generated valid candidate by LSBert. In this section, we will analyze the 8 sentences on Table \ref{sg_example}.

We can see that LSBert can generate one or two valid substitute candidates on these sentences (sent4, sent5, sent7 and sent8), e.g, "senior-$>$powerful", "full-fledged-$>$development", "kinetic-$>$dynamic", and "edited-$>$altered". Since the labels are provided by humans, it is impossible to provide all suitable substitutes in labels. LSBert fail to produce any valid candidate word on the other sentences. When we analyze these wrong substitute candidates, we can find that they can fit the context. We can guess that LSBert mainly focuses more on the context and ignores the meaning of the original word on these wrong examples.

\textbf{ (2) The analysis of substitute ranking results}

\begin{table*}
\centering
\begin{tabular}{l|l}
\hline
Sent1 & Triangles can also be classified according to their internal angles, measured here in degrees. \\

Labels & grouped, categorized, arranged, labeled, divided, organized, separated, defined, described, ...\\

LSBert\_SR &  \textbf{divided}, \textbf{described}, separated, designated, ...\\
LSBert\_Substitute & classified \\
\hline
Sent2 & ...; he retained the conductorship of the Vienna Philharmonic until 1927. \\

Labels & kept, held, had, got\\

LSBert\_SR & maintained, \textbf{held}, \textbf{kept}, remained, continued, shared, ... \\
LSBert\_Substitute & maintained \\
\hline

Sent3 & ..., and a Venetian in Paris in 1528 also reported that she was said to be beautiful \\

Labels & said, told, stated, wrote, declared, indicated, noted, claimed, announced, mentioned \\

LSBert\_SR & \textbf{noted}, confirmed, described, claimed, recorded, \textbf{said}, ... \\
LSBert\_Substitute & reported \\
\hline

Sent4 & ..., the king will rarely play an active role in the development of an offensive or .... \\

Labels & infrequently, hardly, uncommonly, barely, seldom, unlikely, sometimes, not, seldomly, ...\\

LSBert\_SR & never, usually, \textbf{seldom}, \textbf{not}, \textbf{barely}, \textbf{hardly}, ... \\
LSBert\_Substitute & never \\
\hline
\end{tabular}
\caption{The examples that the final substitute generated by LSBert is not from the labels. The words in the substitute ranking belonging to the labels are shown in bold.}
\label{sr_example}
\end{table*}

LSBert can find one or more suitable alternatives for almost all samples, but the final system results do not always select the most suitable candidate as the final substitute. In this section, we will analyze the possible reasons for this question. In Table \ref{sr_example}, we give some examples that LSBert cannot produce the right substitute ranking. 

From sent1 and sent3, we can see that the substitute ranking (SR) chooses the best substitute, but LSBert still chooses the original word. This is because the Zipf value of "divided" is 3.65 and the Zipf value of "classified" is 3.83, LSBert considers "classified" to be simpler than "divided". It is the same reason for sent3 in which the Zipf value of "noted" is 3.68 and the Zipf value of "reported" is 4.18. Consequently, in sent1 and sent3, the best substitutes of SR cannot be used as the final substitutes.

The second case is that the best substitution of the SR step is not from the labels provided by humans. In sent2 and sent4, LSBert chooses "maintained"  as a simpler for "retained" and "never" as a simpler for "rarely". We can find that these words "maintained" and "never" are also suitable substitutes, but do not appear in the labels.

\textbf{ (3) The analysis of sentence simplification results}

\begin{table*}
\centering
\begin{tabular}{|c|l|l|}
\hline
\multirow{5}{*}{1}  & Sentence &  Admission to Tsinghua is exceedingly competitive. \\
 & Label &  Entrance to Tsinghua is very very difficult. \\
\hdashline

& Glava{\v{s}} & \textbf{Offers} to \textbf{Qinghua} is \textbf{very} \textbf{exciting}. \\

& REC-LS & Admission to Tsinghua is exceedingly competitive. \\

& LSBert & \textbf{Entrance} to Tsinghua is \textbf{very} \textbf{tough}.\\
\hline
\multirow{4}{*}{2} & Sentence &  Many species had vanished by the end of the nineteenth century, with European settlement. \\
& Label &  With Euopean settlement many species have been vanished. \\
\hdashline

& Glava{\v{s}} & \textbf{Some} \textbf{birds} was \textbf{gone} by the \textbf{time} of the \textbf{twentieth} \textbf{history}, with \textbf{world} \textbf{land}. \\

& REC-LS  &  Many species had \textbf{disappeared} by the end of the \textbf{19th} century, with European settlement. \\

& LSBert & Many \textbf{animals} had \textbf{disappeared} by the end of the nineteenth century, with European settlement. \\
\hline
\multirow{4}{*}{3} & Sentence &  In 1987 Wexler was inducted into the Rock and Roll Hall of Fame. \\
& Label &  In 1987 Wexler was inducted into the Rock and Roll Hall of Fame.  \\
\hdashline

& Glava{\v{s}} & In 1987 \textbf{Livingston} was \textbf{fame} into the rock and \textbf{you} hall of \textbf{hall}. \\
& REC-LS & In 1987 Wexler was inducted into the Rock and Roll Hall of Fame. \\
& LSBert & In 1987 Wexler was \textbf{elected} into the Rock and Roll Hall of \textbf{Honor}.\\
\hline
\multirow{4}{*}{4} & Sentence &  Oregano is an indispensable ingredient in Greek cuisine. \\
 & Label & Oregano is a necessary ingredient in Greek cuisine.\\
\hdashline
& Glava{\v{s}} & \textbf{Garlic} is an \textbf{essential} ingredient in Greek \textbf{cooking}.  \\
& REC-LS & Oregano is an \textbf{essential} \textbf{element} in Greek cuisine.  \\
& LSBert & Oregano is an \textbf{important} \textbf{element} in Greek \textbf{food}. \\
\hline
\multirow{4}{*}{5} & Sentence &  Their eyes are quite small, and their visual acuity is poor. \\
& Label &  Their eyes are quite small, and their visual acuity is poor. \\
\hdashline

& Glava{\v{s}} & Their eyes \textbf{have} \textbf{very} \textbf{little}, and their \textbf{musical} \textbf{visual} is \textbf{bad}.  \\
& REC-LS & Their eyes are quite small, and their \textbf{ocular} \textbf{acuteness} is poor.\\
& LSBert &  Their eyes are quite small, and their visual \textbf{ability} is \textbf{bad}. \\
\hline
\end{tabular}
\caption{The simplified sentences are shown using three different LS methods on WikiLarge dataset. Substitutions are shown in bold. }
\label{TSExample}
\end{table*}

The above qualitative study for LSBert need to provide the complex word by humans. In this experiment, we try to verify the results of LS methods on sentence simplification. We also choose the two methods Glava{\v{s}} and REC-LS to comparison. Table \ref{TSExample} shows some examples from the WikiLarge dataset to be simplified. We note that we draw the same conclusions from these examples with LS system for sentence simplification. Glava{\v{s} tries to simplify every content word in the sentence ignoring the aim of LS. LS aims to replace complex words in a given sentence with simpler alternatives of equivalent meaning. Rec-LS can make the right simplifications, e.g., sentence 2. But, for sentence 1 and sentence 3, Rec-LS outputs the original sentence. LSBert replaces complex words with simpler alternatives and makes the most reasonable simplification. This verifies that our framework LSBert fits for lexical simplification.

\setlength\dashlinedash{1.5pt}

\section{Conclusion} 

We propose a simple BERT-based framework LSBert for lexical simplification (LS) by leveraging the idea of masking language model of Bert. The existing LS methods only consider the context of the complex word on the last step (substitute ranking) of LS. LSBert focuses on the context of the complex word on all steps of lexical simplification without relying on the parallel corpus or linguistic databases. Experiment results have shown that our approach LSBert achieves the best performance on three well-known benchmarks. Since Bert can be trained in raw text, our method can be applied to many languages for lexical simplification. One limitation of our method is that it only generates a single-word replacement for the complex word, but we plan to extend it to support multi-word expressions. In the future, the pretrained Bert model can be fine-tuned with just simple English corpus (e.g., Newsela), and then we will use fine-tuned Bert for lexical simplification.

\section*{Acknowledgement}

This research is partially supported by the National Natural Science Foundation of China under grants 61703362 and 91746209; the National Key Research and Development Program of China under grant 2016YFB1000900; the Program for Changjiang Scholars and Innovative Research Team in University (PCSIRT) of the Ministry of Education, China, under grant IRT17R32; and the Natural Science Foundation of Jiangsu Province of China under grant BK20170513. This manuscript is an extended version of the conference paper, titled “Lexical Simplification with Pretrained Encoders”, published in the Thirty-Fourth AAAI Conference on Artificial Intelligence (AAAI), New York, February 7-12, 2020.

\bibliographystyle{IEEEtran}
\bibliography{BERTLS}

\begin{thebibliography}{10}
\expandafter\ifx\csname url\endcsname\relax
  \def\url#1{\texttt{#1}}\fi
\expandafter\ifx\csname urlprefix\endcsname\relax\def\urlprefix{URL }\fi
\expandafter\ifx\csname href\endcsname\relax
  \def\href#1#2{#2} \def\path#1{#1}\fi

\bibitem{De_belder}
J.~De~Belder, M.-F. Moens, Text simplification for children, In Proceedings of
  the 2010 SIGIR Workshop on Accessible Search Systems (2010) 19--26.

\bibitem{paetzold2016unsupervised}
G.~H. Paetzold, L.~Specia, Unsupervised lexical simplification for non-native
  speakers., in: AAAI, 2016, pp. 3761--3767.

\bibitem{feng2009automatic}
L.~Feng, Automatic readability assessment for people with intellectual
  disabilities, ACM SIGACCESS accessibility and computing~(93) (2009) 84--91.

\bibitem{saggion2017automatic}
H.~Saggion, Automatic text simplification, Synthesis Lectures on Human Language
  Technologies 10~(1) (2017) 1--137.

\bibitem{paetzold2017lexical}
G.~Paetzold, L.~Specia, Lexical simplification with neural ranking, in: ACL:
  Volume 2, Short Papers, 2017, pp. 34--40.

\bibitem{gooding2019recursive}
S.~Gooding, E.~Kochmar, Recursive context-aware lexical simplification, in:
  Proceedings of the 2019 Conference on Empirical Methods in Natural Language
  Processing and the 9th International Joint Conference on Natural Language
  Processing (EMNLP-IJCNLP), 2019, pp. 4855--4865.

\bibitem{devlin1998the}
S.~{Devlin}, J.~{Tait}, The use of a psycholinguistic database in the simpli
  cation of text for aphasic readers, Linguistic Databases 1 (1998) 161–173.

\bibitem{pavlick2016simple}
E.~Pavlick, C.~Callison-Burch, Simple ppdb: A paraphrase database for
  simplification, in: ACL: Volume 2, Short Papers, 2016, pp. 143--148.

\bibitem{glavavs2015simplifying}
G.~Glava{\v{s}}, S.~{\v{S}}tajner, Simplifying lexical simplification: do we
  need simplified corpora?, in: ACL, 2015, pp. 63--68.

\bibitem{gooding2019complex}
S.~Gooding, E.~Kochmar, Complex word identification as a sequence labelling
  task, in: Proceedings of the 57th Annual Meeting of the Association for
  Computational Linguistics, 2019, pp. 1148--1153.

\bibitem{devlin2018bert}
J.~Devlin, M.-W. Chang, K.~Lee, K.~Toutanova, Bert: Pre-training of deep
  bidirectional transformers for language understanding, arXiv preprint
  arXiv:1810.04805.

\bibitem{coster2011simple}
W.~Coster, D.~Kauchak, Simple english wikipedia: a new text simplification
  task, in: ACL, 2011, pp. 665--669.

\bibitem{wang2016experimental}
T.~Wang, P.~Chen, K.~Amaral, J.~Qiang, An experimental study of lstm
  encoder-decoder model for text simplification, arXiv preprint
  arXiv:1609.03663.

\bibitem{nisioi2017exploring}
S.~Nisioi, S.~{\v{S}}tajner, S.~P. Ponzetto, L.~P. Dinu, Exploring neural text
  simplification models, in: ACL, Vol.~2, 2017, pp. 85--91.

\bibitem{dong2019editnts}
Y.~Dong, Z.~Li, M.~Rezagholizadeh, J.~C.~K. Cheung, Editnts: An neural
  programmer-interpreter model for sentence simplification through explicit
  editing, in: Proceedings of the 57th Annual Meeting of the Association for
  Computational Linguistics, 2019, pp. 3393--3402.

\bibitem{zhu2010monolingual}
Z.~Zhu, D.~Bernhard, I.~Gurevych, A monolingual tree-based translation model
  for sentence simplification, in: Proceedings of the 23rd international
  conference on computational linguistics, 2010, pp. 1353--1361.

\bibitem{zhang2017sentence}
X.~Zhang, M.~Lapata, Sentence simplification with deep reinforcement learning,
  arXiv preprint arXiv:1703.10931.

\bibitem{xu2015problems}
W.~Xu, C.~Callison-Burch, C.~Napoles, Problems in current text simplification
  research: New data can help, TACL 3~(1) (2015) 283--297.

\bibitem{vstajner2015deeper}
S.~{\v{S}}tajner, H.~B{\'e}chara, H.~Saggion, A deeper exploration of the
  standard pb-smt approach to text simplification and its evaluation, in: ACL,
  2015, pp. 823--828.

\bibitem{hwang2015aligning}
W.~Hwang, H.~Hajishirzi, M.~Ostendorf, W.~Wu, Aligning sentences from standard
  wikipedia to simple wikipedia, in: ACL, 2015, pp. 211--217.

\bibitem{shardlow2014survey}
M.~Shardlow, A survey of automated text simplification, International Journal
  of Advanced Computer Science and Applications 4~(1) (2014) 58--70.

\bibitem{paetzold2017survey}
G.~H. Paetzold, L.~Specia, A survey on lexical simplification, in: Journal of
  Artificial Intelligence Research, Vol.~60, 2017, pp. 549--593.

\bibitem{Lesk:1986:ASD:318723.318728}
M.~Lesk, Automatic sense disambiguation using machine readable dictionaries:
  How to tell a pine cone from an ice cream cone, in: Proceedings of the 5th
  Annual International Conference on Systems Documentation, 1986, pp. 24--26.

\bibitem{maddela-xu-2018-word}
M.~Maddela, W.~Xu, A word-complexity lexicon and a neural readability ranking
  model for lexical simplification, in: EMNLP, 2018, pp. 3749--3760.

\bibitem{biran2011putting}
O.~Biran, S.~Brody, N.~Elhadad, Putting it simply: a context-aware approach to
  lexical simplification, in: ACL, 2011, pp. 496--501.

\bibitem{yatskar2010sake}
M.~Yatskar, B.~Pang, C.~Danescu-Niculescu-Mizil, L.~Lee, For the sake of
  simplicity: Unsupervised extraction of lexical simplifications from
  wikipedia, in: NACACL, Association for Computational Linguistics, 2010, pp.
  365--368.

\bibitem{horn2014learning}
C.~Horn, C.~Manduca, D.~Kauchak, Learning a lexical simplifier using wikipedia,
  in: ACL, Volume 2: Short Papers, 2014, pp. 458--463.

\bibitem{lee2019biobert}
J.~Lee, W.~Yoon, S.~Kim, D.~Kim, S.~Kim, C.~H. So, J.~Kang, Biobert:
  pre-trained biomedical language representation model for biomedical text
  mining, arXiv preprint arXiv:1901.08746.

\bibitem{lample2019cross}
G.~Lample, A.~Conneau, Cross-lingual language model pretraining, arXiv preprint
  arXiv:1901.07291.

\bibitem{qiang2020AAAI}
J.~Qiang, Y.~Li, Y.~Zhu, Y.~Yuan, X.~Wu, Lexical simplification with pretrained
  encoders, Thirty-Fourth AAAI Conference on Artificial Intelligence.

\bibitem{zhou2019bert}
W.~Zhou, T.~Ge, K.~Xu, F.~Wei, M.~Zhou, Bert-based lexical substitution, in:
  Proceedings of the 57th Annual Meeting of the Association for Computational
  Linguistics, 2019, pp. 3368--3373.

\bibitem{shardlow2013comparison}
M.~Shardlow, A comparison of techniques to automatically identify complex
  words., in: 51st Annual Meeting of the Association for Computational
  Linguistics Proceedings of the Student Research Workshop, 2013, pp. 103--109.

\bibitem{yimam2018report}
S.~M. Yimam, C.~Biemann, S.~Malmasi, G.~H. Paetzold, L.~Specia,
  S.~{\v{S}}tajner, A.~Tack, M.~Zampieri, A report on the complex word
  identification shared task 2018 (2018) 66--78.

\bibitem{yimam2017cwig3g2}
S.~M. Yimam, S.~{\v{S}}tajner, M.~Riedl, C.~Biemann, Cwig3g2-complex word
  identification task across three text genres and two user groups, in:
  Proceedings of the Eighth International Joint Conference on Natural Language
  Processing (Volume 2: Short Papers), 2017, pp. 401--407.

\bibitem{Brysbaert2009Moving}
M.~Brysbaert, B.~New, Moving beyond kucera and francis: A critical evaluation
  of current word frequency norms and the introduction of a new and improved
  word frequency measure for american english, Behavior Research Methods 41~(4)
  (2009) 977--990.

\bibitem{Kriz2018Simplification}
R.~Kriz, E.~Miltsakaki, M.~Apidianaki, C.~Callisonburch, Simplification using
  paraphrases and context-based lexical substitution, in: Conference of the
  North American Chapter of the Association for Computational Linguistics:
  Human Language Technologies, 2018, pp. 207--217.

\bibitem{Ganitkevitch2013}
J.~Ganitkevitch, B.~V. Durme, C.~Callison-Burch, Ppdb: The paraphrase database,
  in: NAACL-HLT, 2013, pp. 758--764.

\bibitem{kajiwara2013selecting}
T.~Kajiwara, H.~Matsumoto, K.~Yamamoto, Selecting proper lexical paraphrase for
  children, in: ROCLING, 2013, pp. 59--73.

\bibitem{paetzold2015lexenstein}
G.~Paetzold, L.~Specia, Lexenstein: A framework for lexical simplification, in:
  Proceedings of ACL-IJCNLP 2015 System Demonstrations, 2015, pp. 85--90.

\bibitem{Jipeng2019Unsupervised}
J.~Qiang, X.~Wu, Unsupervised statistical text simplification, IEEE
  Transactions on Knowledge and Data Engineering.

\bibitem{martin2019controllable}
L.~Martin, B.~Sagot, {\'E}.~de~la Clergerie, A.~Bordes, Controllable sentence
  simplification, arXiv preprint arXiv:1910.02677.

\bibitem{woodsend2011learning}
K.~Woodsend, M.~Lapata, Learning to simplify sentences with quasi-synchronous
  grammar and integer programming, in: EMNLP, 2011, pp. 409--420.

\bibitem{xu2016optimizing}
W.~Xu, C.~Napoles, E.~Pavlick, Q.~Chen, C.~Callison-Burch, Optimizing
  statistical machine translation for text simplification, TACL 4 (2016)
  401--415.

\bibitem{kincaid1975}
P.~J. Kincaid, R.~P. Fishburne, R.~J.~L. Richard, B.~S. Chissom, Derivation of
  new readability formulas (automated readability index, fog count and flesch
  reading ease formula) for navy enlisted personnel, Technical report, DTIC
  Document.

\end{thebibliography}

\begin{IEEEbiography}[{\includegraphics[width=1in,height=1.25in,clip,keepaspectratio]{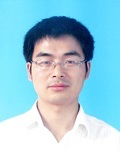}}]%
{Jipeng Qiang} is an assistant professor and the group leader of Computational Linguistics and Data Mining Group at Yanghou University. He received his Ph.D. degree in computer science and technology from Hefei university of Technology in 2016. He was a Ph.D. visiting student in Artificial Intelligence Lab at the University of Massachusetts Boston from 2014 to 2016. His research interests mainly include data mining and computational linguistics. He has received one grant from National Natural Science Foundation of China, one grant from Natural Science Foundation of Jiangsu Province of China, one grant from Natural Science Foundation of the Higher Education Institutions of Jiangsu Province of China. He has published more than 40 papers in data mining, artificial intelligence, and computational linguistics conferences and journals. \end{IEEEbiography}

\begin{IEEEbiography}[{\includegraphics[width=1in,height=1.25in,clip,keepaspectratio]{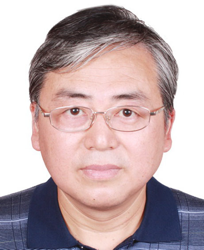}}]%
{Yun Li} is currently a professor in the School of Information Engineering, Yangzhou University, China. He received the M.S. degree in computer science and technology from Hefei University of Technology, China, in 1991, and the Ph.D. degree in control theory and control engineering from Shanghai University, China, in 2005. He has published more than 100 scientific papers. His research interests include data mining and cloud computing. 
\end{IEEEbiography}

\begin{IEEEbiography}[{\includegraphics[width=1in,height=1.25in,clip,keepaspectratio]{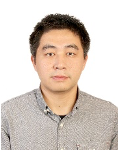}}]%
{Yi Zhu} is currently an assistant professor in the School of information Engineering, Yangzhou University, China. He received the BS degree from Anhui University, the MS degree from University of Science and Technology of China, and the PhD degree from Hefei University of Technology. His research interests are in data mining and knowledge engineering. His research interests include data mining, knowledge engineering, and recommendation systems.. \end{IEEEbiography}

\begin{IEEEbiography}[{\includegraphics[width=1in,height=1.25in,clip,keepaspectratio]{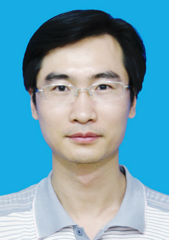}}]%
{Yunhao Yuan} is currently an associate professor in the School of information Engineering, Yangzhou University, China. He received the M. Eng. degree in computer science and technology from Yangzhou University, China, in 2009, and the Ph.D. degree in pattern recognition and intelligence system from Nanjing University of Science and Technology, China, in 2013. His research interests include pattern recognition, data mining, and image processing. 
\end{IEEEbiography}

\begin{IEEEbiography}[{\includegraphics[width=1in,height=1.25in,clip,keepaspectratio]{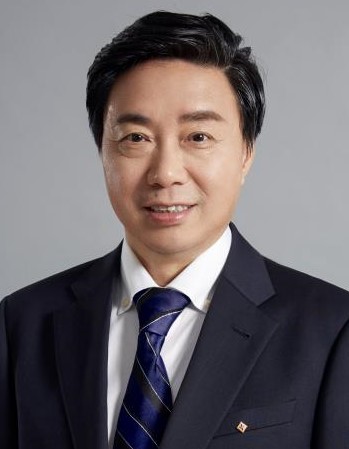}}]{Xindong Wu} is a Yangtze River Scholar in the School of Computer Science and Information Engineering at the Hefei University of Technology, China, and the president of Mininglamp Academy of Sciences, Minininglamp, Beijing, China, and a fellow of IEEE and AAAS. He received his B.S. and M.S. degrees in computer science from the Hefei University of Technology, China, and his Ph.D. degree in artificial intelligence from the University of Edinburgh, Britain. His research interests include data mining, big data analytics, knowledge-based systems, and Web information exploration. He is currently the steering committee chair of the IEEE International Conference on Data Mining (ICDM), the editor-in-chief of Knowledge and Information Systems (KAIS, by Springer), and a series editor-in-chief of the Springer Book Series on Advanced Information and Knowledge Processing (AI\&KP). He was the editor-in-chief of the IEEE Transactions on Knowledge and Data Engineering (TKDE, by the IEEE Computer Society) between 2005 and 2008. He served as program committee chair/co-chair for the 2003 IEEE International Conference on Data Mining, the 13th ACM SIGKDD International Conference on Knowledge Discovery and Data Mining, and the 19th ACM Conference on Information and Knowledge Management.
\end{IEEEbiography}

\end{document}